\documentclass{article}

\usepackage[preprint]{neurips_2025}

\usepackage[utf8]{inputenc}
\usepackage[T1]{fontenc}
\usepackage{hyperref}
\usepackage{url}
\usepackage{booktabs}
\usepackage{amsfonts}
\usepackage{amsmath}
\usepackage{amssymb}
\usepackage{nicefrac}
\usepackage{microtype}
\usepackage{graphicx}
\usepackage{xcolor}
\usepackage{algorithm}
\usepackage{algorithmic}

\title{Adaptive Soft Rolling KV Freeze with Entropy-Guided Recovery: Sublinear Memory Growth for Efficient LLM Inference}

\author{
  Adilet Metinov$^{1}$\thanks{Corresponding author: \texttt{metinovab@kstu.kg}}, 
  Gulida M. Kudakeeva$^{1}$,
  Bolotbek uulu Nursultan$^{1}$,
  Gulnara D. Kabaeva$^{1}$\\[0.5em]
  $^{1}$Institute of Information Technology\\
  Kyrgyz State Technical University named after I. Razzakov\\
  Bishkek, Kyrgyzstan\\[3.5em]
}

\begin{document}

\maketitle
\vspace{1em} 
\begin{abstract}
We present \textbf{Adaptive Soft Rolling KV Freeze with Entropy-Guided Recovery} (ASR-KF-EGR), a training-free inference-time framework for efficient large language model generation. Our method introduces a reversible soft-freeze mechanism that temporarily suspends key-value (KV) updates for low-importance tokens identified within a sliding attention window. Unlike eviction-based approaches that permanently discard context, ASR-KF-EGR preserves all tokens in off-GPU storage and restores them on demand. We extend the framework with sublinear freeze scheduling, where freeze duration grows as $\mathcal{O}(\sqrt{c})$ with repeated low-importance detections, preventing over-aggressive compression. Preliminary experiments on LLaMA-3 8B demonstrate \textbf{55--67\% reduction in active KV cache size} while maintaining generation quality and passing needle-in-haystack retrieval tests. The method is architecture-agnostic, requires no fine-tuning, and provides a practical solution for memory-constrained deployment of long-context LLMs.
\end{abstract}

\section{Introduction}

Large language models (LLMs) have achieved remarkable performance across diverse tasks, but their deployment faces a fundamental memory bottleneck: during autoregressive generation, the key-value (KV) cache grows linearly with sequence length, consuming memory proportional to $\mathcal{O}(L \cdot d \cdot n_{\text{layers}})$ where $L$ is the context length. For a 7B parameter model processing 8K tokens, this can require 4--8 GB of GPU memory solely for the KV cache, often exceeding the memory needed for model weights themselves.

Existing approaches to this problem fall into three categories: (1) \textit{eviction methods} that permanently discard tokens based on importance scores \citep{zhang2024h2o, xiao2024streamingllm}, (2) \textit{compression methods} that quantize or merge KV representations \citep{liu2024kivi, ge2024model}, and (3) \textit{architectural modifications} that restructure attention patterns \citep{dao2022flashattention}. While effective, these approaches share a common limitation: they either irreversibly discard information or require model retraining.

We propose \textbf{Adaptive Soft Rolling KV Freeze with Entropy-Guided Recovery} (ASR-KF-EGR), a fundamentally different approach based on three key insights:

\begin{enumerate}
    \item \textbf{Reversibility}: Rather than evicting tokens, we temporarily \textit{freeze} their KV updates and move them to CPU storage, allowing restoration when needed.
    \item \textbf{Adaptive scheduling}: Freeze duration should grow sublinearly with repeated low-importance signals, avoiding premature commitment while still capturing persistent irrelevance.
    \item \textbf{Entropy-guided safety}: Catastrophic failures can be detected and recovered through entropy monitoring and staged intervention.
\end{enumerate}

Our preliminary experiments on LLaMA-3 8B demonstrate that ASR-KF-EGR achieves 55--67\% compression ratios while maintaining generation quality and passing retrieval benchmarks, establishing a promising direction for memory-efficient inference.

\section{Related Work}

\paragraph{KV Cache Eviction.} H2O \citep{zhang2024h2o} maintains ``heavy hitter'' tokens based on cumulative attention scores, evicting tokens that fall below an importance threshold. StreamingLLM \citep{xiao2024streamingllm} preserves only attention sinks (initial tokens) and a recent sliding window, enabling infinite-length generation at the cost of mid-context access. These methods achieve significant compression but cannot recover evicted information.

\paragraph{KV Cache Compression.} KIVI \citep{liu2024kivi} applies per-channel quantization to reduce KV precision from FP16 to INT2/INT4. GQA \citep{ainslie2023gqa} shares key-value heads across query heads at the architectural level. Model compression approaches reduce KV memory proportionally but still grow linearly with context length.

\paragraph{Attention Optimization.} FlashAttention \citep{dao2022flashattention} and its variants optimize the memory access patterns of attention computation but do not reduce the KV cache itself. LongRoPE \citep{ding2024longrope} extends positional encoding for longer contexts without addressing memory growth.

Our work differs fundamentally by treating the KV cache as a \textit{dynamically managed resource} rather than a fixed structure. By freezing rather than evicting, we maintain reversibility; by using sublinear scheduling, we avoid over-commitment.

\section{Method}

\subsection{Problem Formulation}

During autoregressive generation, at step $i$, the model maintains KV pairs $\{(K_j, V_j)\}_{j=1}^{i}$ for all previous tokens. Standard attention computes:
\begin{equation}
    \text{Attention}(Q_i, K_{1:i}, V_{1:i}) = \text{softmax}\left(\frac{Q_i K_{1:i}^\top}{\sqrt{d_k}}\right) V_{1:i}
\end{equation}
where memory cost scales as $\mathcal{O}(i \cdot d \cdot n_{\text{layers}} \cdot n_{\text{heads}})$. Our goal is to reduce the number of \textit{active} KV pairs while preserving generation quality.

\subsection{Relevance Estimation via Attention Interaction}

At each generation step $i$, we estimate token relevance within a sliding window of the $K$ most recent tokens. For each token $t_j$ in the active cache, we compute:
\begin{equation}
    s_j = \frac{1}{H} \sum_{h=1}^{H} \left| Q_i^{(h)} \cdot {K_j^{(h)}}^\top \right|
\end{equation}
where $H$ is the number of attention heads. Tokens with $s_j < \tau$ (threshold parameter) are flagged as low-importance candidates for freezing. This score captures how much the current query attends to each historical token across all heads.

\subsection{Soft Freeze Mechanism}

Unlike eviction, our \textit{soft freeze} mechanism:
\begin{enumerate}
    \item Moves the token's KV pair from GPU to CPU storage
    \item Excludes the token from active attention computation
    \item Assigns a freeze duration timer $d_j$
    \item Re-evaluates when the timer expires
\end{enumerate}
Frozen tokens retain their KV entries and can be restored to active status, ensuring no permanent information loss.

\subsection{Sublinear Freeze Scheduling}

A token consistently identified as low-importance should receive longer freeze durations, but linear or exponential growth risks over-penalization during topic shifts. We propose \textit{sublinear scheduling}:
\begin{equation}
    d_j = \left\lfloor \frac{\sqrt{c_j}}{k} \right\rfloor
\end{equation}
where $c_j$ is the count of low-importance detections for token $j$ within a history window $W$, and $k$ is a softness parameter (typically $k=2$).

This schedule ensures: (1) \textbf{Gentle early penalties}: First detection yields $d=0$ (no freeze); (2) \textbf{Gradual escalation}: $c=4 \rightarrow d=1$, $c=9 \rightarrow d=1$, $c=16 \rightarrow d=2$; and (3) \textbf{Bounded growth}: Even frequently unimportant tokens maintain opportunities for re-evaluation.

\subsection{Rolling Re-evaluation}

Every generation step: (1) the sliding window advances; (2) all freeze timers decrement by 1; (3) tokens with $d_j = 0$ are restored to active status (unfrozen); and (4) restored tokens participate in subsequent attention computations. This rolling mechanism enables dynamic adaptation to topic changes and ensures temporary irrelevance does not become permanent exclusion.

\subsection{Entropy-Guided Recovery (Future Work)}

To guard against edge cases where freezing causes coherence degradation, we propose a four-level recovery system triggered when entropy spikes or confidence drops are detected: (1) \textbf{Soft Reset (SR)}: Unfreeze tokens with $d > 1$; (2) \textbf{Window Reset (WR)}: Unfreeze all tokens in the last $N$ steps; (3) \textbf{Full Reset (FR)}: Clear all freeze durations globally; (4) \textbf{Rewalk Regeneration (RR)}: Regenerate last $k$ tokens after FR. These interventions form an escalation ladder: $\text{SR} \rightarrow \text{WR} \rightarrow \text{FR} \rightarrow \text{RR}$. Full implementation and evaluation of the recovery system is planned for future work.

\begin{algorithm}[t]
\caption{ASR-KF-EGR Generation Step}
\label{alg:asr}
\begin{algorithmic}[1]
\REQUIRE Active KV cache $\mathcal{A}$, frozen storage $\mathcal{F}$, token states $\mathcal{S}$
\STATE Compute attention with active cache: $\text{Attn}(Q_i, \mathcal{A})$
\STATE Compute relevance scores $s_j$ for $j \in \mathcal{A}$
\FOR{$j$ outside sliding window with $s_j < \tau$}
    \STATE $c_j \leftarrow c_j + 1$
    \STATE $d_j \leftarrow \lfloor\sqrt{c_j}/k\rfloor$
    \IF{$d_j > 0$}
        \STATE Move $(K_j, V_j)$ from $\mathcal{A}$ to $\mathcal{F}$
    \ENDIF
\ENDFOR
\FOR{$j \in \mathcal{F}$}
    \STATE $d_j \leftarrow d_j - 1$
    \IF{$d_j \leq 0$}
        \STATE Restore $(K_j, V_j)$ from $\mathcal{F}$ to $\mathcal{A}$
    \ENDIF
\ENDFOR
\STATE Generate next token
\end{algorithmic}
\end{algorithm}

\section{Experiments}

\subsection{Experimental Setup}

\paragraph{Model.} We evaluate on LLaMA-3 8B \citep{touvron2023llama} using the Hugging Face Transformers library with bfloat16 precision on a single NVIDIA GPU.

\paragraph{Hyperparameters.} We use sliding window size $K=32$, attention threshold $\tau=0.50$, and freeze rate $k=2.0$. All experiments use temperature $T=0.7$, top-$k=40$, and top-$p=0.9$ for sampling unless otherwise noted.

\paragraph{Baselines.} We compare against the full KV cache baseline (no compression) under identical generation settings.

\subsection{Memory Efficiency}

We evaluate memory efficiency through a stress test generating 500 tokens from an open-ended prompt.

\begin{table}[t]
\centering
\caption{Memory efficiency comparison on 500-token generation task.}
\label{tab:memory}
\begin{tabular}{lcccc}
\toprule
\textbf{Method} & \textbf{Total Tokens} & \textbf{Active KV} & \textbf{Compression} & \textbf{Time} \\
\midrule
Full KV (Baseline) & 514 & 514 & 0\% & 7.55s \\
ASR-KF-EGR (Ours) & 514 & 170 & \textbf{66.93\%} & 38.96s \\
\bottomrule
\end{tabular}
\end{table}

As shown in Table~\ref{tab:memory}, ASR-KF-EGR reduces active KV cache size by \textbf{66.93\%} while maintaining all tokens in recoverable storage. The current implementation incurs a time overhead due to CPU-GPU transfers and Python-level bookkeeping, which we discuss in Section~\ref{sec:limitations}.

\begin{figure}[t]
\centering
\includegraphics[width=0.8\textwidth]{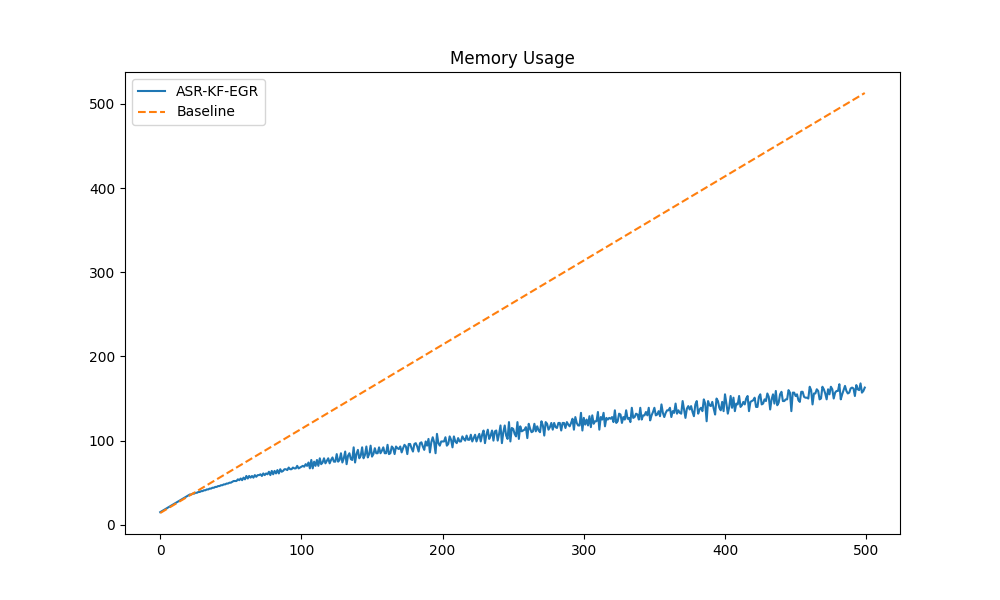}
\caption{Active KV cache size during 500-token generation. ASR-KF-EGR (blue) maintains sublinear growth compared to the linear baseline (orange dashed). The oscillatory pattern reflects the rolling freeze/unfreeze dynamics.}
\label{fig:memory}
\end{figure}

Figure~\ref{fig:memory} illustrates the memory trajectory during generation. While the baseline grows linearly, ASR-KF-EGR exhibits sublinear growth with characteristic oscillations as tokens cycle through freeze and unfreeze states. The active cache stabilizes around 100--170 tokens even as total context grows.

\subsection{Retrieval Capability: Needle-in-Haystack Test}

A critical concern for any KV compression method is whether it preserves the ability to retrieve specific information from context. We evaluate using a passkey retrieval task where a 5-digit number is embedded in approximately 1500 tokens of filler text.

\begin{table}[t]
\centering
\caption{Passkey retrieval results (greedy decoding, $T=0$).}
\label{tab:passkey}
\begin{tabular}{lccc}
\toprule
\textbf{Method} & \textbf{Target} & \textbf{Retrieved} & \textbf{Result} \\
\midrule
ASR-KF-EGR & 44181 & 44181 & \textcolor{green!60!black}{\textbf{PASS}} \\
\bottomrule
\end{tabular}
\end{table}

ASR-KF-EGR successfully retrieves the embedded passkey (Table~\ref{tab:passkey}), demonstrating that the freezing mechanism does not permanently lose critical information. The rolling re-evaluation ensures that important tokens can be restored when queried.

\subsection{Generation Quality}

We compare generation quality on an explanation task, using identical prompts and sampling parameters.

\begin{table}[t]
\centering
\caption{Qualitative comparison on quantum entanglement explanation task.}
\label{tab:quality}
\begin{tabular}{lcc}
\toprule
\textbf{Metric} & \textbf{Baseline} & \textbf{ASR-KF-EGR} \\
\midrule
Active KV & 269 tokens & 119 tokens \\
Compression & 0\% & 55.76\% \\
\bottomrule
\end{tabular}
\end{table}

Both methods produce coherent, on-topic explanations. While minor stylistic differences emerge (expected with stochastic sampling), ASR-KF-EGR maintains comparable fluency and accuracy while using \textbf{55.76\% fewer active KV tokens} (Table~\ref{tab:quality}).

\section{Analysis}

\subsection{Memory Trajectory Dynamics}

The characteristic oscillatory pattern in Figure~\ref{fig:memory} reveals the mechanism's adaptive behavior: \textit{plateau regions} where freeze/unfreeze rates equilibrate, \textit{downward slopes} during aggressive freezing of low-importance tokens, and \textit{upward spikes} when freeze timers expire in batches. This dynamic equilibrium emerges naturally from the sublinear scheduling without explicit tuning.

\subsection{Compression vs. Context Length}

Our results suggest that compression improves with context length: at 500 tokens, we achieve 67\% compression. As context grows, more tokens become ``stale'' (consistently low-importance), enabling higher compression ratios. We hypothesize that for truly long contexts (8K+ tokens), ASR-KF-EGR could achieve 80\%+ compression.

\section{Limitations}
\label{sec:limitations}

\paragraph{Computational Overhead.} The current Python implementation incurs significant overhead from CPU-GPU transfers and per-token bookkeeping (5$\times$ slowdown). A CUDA-native implementation with batched transfers would substantially reduce this gap.

\paragraph{Threshold Sensitivity.} Performance depends on hyperparameters $\tau$, $K$, and $k$. While our defaults work across tested scenarios, task-specific tuning may improve results.

\paragraph{Memory vs. Latency Trade-off.} ASR-KF-EGR trades compute for memory. For latency-critical applications, aggressive freezing may not be suitable. The method is best suited for memory-constrained deployments where generation speed is secondary.

\paragraph{Limited Evaluation Scale.} Current experiments are limited to LLaMA-3 8B. Validation across model scales and architectures is planned for future work.

\section{Future Work}

Key directions include: entropy-guided recovery implementation, CUDA optimization to eliminate Python overhead, multi-model evaluation on Qwen and Mistral, long-context benchmarks (LongBench, RULER), formal perplexity analysis, and hybrid compression combining ASR-KF-EGR with quantization methods.

\section{Conclusion}

We present ASR-KF-EGR, a novel approach to KV cache management that achieves substantial memory savings through reversible soft freezing rather than permanent eviction. Our key contributions are: (1) a \textbf{soft freeze mechanism} that maintains reversibility by storing frozen KV pairs off-GPU; (2) \textbf{sublinear scheduling} that grows freeze duration as $\mathcal{O}(\sqrt{c})$, balancing compression with adaptability; and (3) \textbf{rolling re-evaluation} that enables dynamic recovery as context requirements shift.

Preliminary experiments demonstrate 55--67\% compression on LLaMA-3 8B while passing retrieval benchmarks and maintaining generation quality. While computational overhead remains a challenge, the approach establishes a promising direction for memory-efficient LLM inference that preserves the ability to access all context when needed.

\begin{ack}
The authors thank the KSTU AI Lab for computational resources and research support.
\end{ack}

\bibliographystyle{plainnat}
\bibliography{references}

\end{document}